\documentclass{article}
\usepackage{spconf,amsmath,graphicx}

\usepackage{amssymb}
\usepackage{cite,url,bm,multirow}
\usepackage[bbgreekl]{mathbbol}
\DeclareMathSymbol{\bbepsilon}{\mathord}{bbold}{"0F}
\newcommand{\argmax}{\mathop{\rm arg~max}\limits}
\newcommand{\MAX}{\mathop{\rm max}\limits}
\newcommand{\BiggParen}[1]{\Biggl\{#1\Biggr\}}
\usepackage{breqn}
\usepackage{widetable}
\usepackage{pifont}
\usepackage{booktabs}
\title{Zero-Shot Pseudo Labels Generation Using SAM and CLIP for Semi-Supervised Semantic Segmentation}
\name{Nagito Saito, Shintaro Ito, Koichi Ito, and Takafumi Aoki}
\address{Graduate School of Information Sciences, Tohoku University, Japan.}
\begin{document}
\ninept
\maketitle
\begin{abstract}
  Semantic segmentation is a fundamental task in medical image analysis and autonomous driving and has a problem with the high cost of annotating the labels required in training.
  To address this problem, semantic segmentation methods based on semi-supervised learning with a small number of labeled data have been proposed.
  For example, one approach is to train a semantic segmentation model using images with annotated labels and pseudo labels.
  In this approach, the accuracy of the semantic segmentation model depends on the quality of the pseudo labels, and the quality of the pseudo labels depends on the performance of the model to be trained and the amount of data with annotated labels.
  In this paper, we generate pseudo labels using zero-shot annotation with the Segment Anything Model (SAM) and Contrastive Language-Image Pretraining (CLIP), improve the accuracy of the pseudo labels using the Unified Dual-Stream Perturbations Approach (UniMatch), and use them as enhanced labels to train a semantic segmentation model.
  The effectiveness of the proposed method is demonstrated through the experiments using the public datasets: PASCAL and MS COCO.
\end{abstract}
\section{Introduction}
\label{sec:intro}

Semantic segmentation is one of the fundamental techniques in computer vision \cite{Minaee-TPAMI-2022}, and has been applied to medical image diagnosis \cite{Ronneberger-MICCAI-2015} and autonomous driving \cite{Pohlen-CVPR-2017}.
Training data with class labels assigned to each pixel is indispensable in semantic segmentation using deep learning.
Although a huge amount of training data is required to perform segmentation with high accuracy, pixel-wise annotation of a huge number of images takes considerable time and effort.

There are three major approaches to training deep learning models: supervised learning, self-supervised learning, and semi-supervised learning.
Supervised learning trains a model using only labeled images.
The quantity of labeled images and the quality of labels have a large impact on accuracy in image segmentation.
The accuracy of this approach is high, while the cost of annotation of the training data is also high.
Self-supervised learning trains models using only unlabeled images.
Although this approach does not require label annotation, it can only be used for tasks where labels can be assigned automatically.
This approach is generally used for pre-training of backbone models.
Semi-supervised learning trains models using labeled and unlabeled images.
Compared to supervised learning, semi-supervised learning allows training with a smaller number of labeled images and is not task-specific unlike self-supervised learning.
Therefore, in this paper, we focus on semantic segmentation using semi-supervised learning.

Semantic segmentation using semi-supervised learning has developed from an adversarial learning framework based on Generative Adversarial Networks (GAN) \cite{Souly-ICCV-2017,Mittal-TPAMI-2021} to a consistency regularization framework \cite{Chen-CVPR-2021,Yang-CVPR-2023}.
In the adversarial learning framework, GAN is used to generate images while adding labels \cite{Souly-ICCV-2017}, or to add labels to images while using knowledge distillation to improve the accuracy of semantic segmentation \cite{Mittal-TPAMI-2021}.
These methods have the problem that the training of GAN becomes unstable when the number of labeled images is small.
In the consistency regularization framework, the model is trained so that the predictions of the model when perturbations are applied to unlabeled images correspond to the predictions of the model when perturbations are not applied.
Cross Pseudo Supervision (CPS) \cite{Chen-CVPR-2021} consists of mini batches of labeled and unlabeled data, and trains the two models to match their predictions.
UniMatch \cite{Yang-CVPR-2023} has been proposed to improve the accuracy of CPS.
UniMatch generates pseudo labels from the predictions when applying weak perturbations to images, therefore, the quality of the pseudo labels has a strong impact on model training.

In this paper, we propose a semantic segmentation method using semi-supervised learning, in which the quality of pseudo labels is improved without the consistency regularization framework.
Pseudo labels are assigned to images based on zero-shot annotation using the Segment Anything Model (SAM) \cite{Kirillov-ICCV-2023}, which is a fundamental model for image segmentation, and the Contrastive Language-Image Pretraining (CLIP) \cite{Radford-ICML-2021}, which is a fundamental model for Vision and Language.
Inspired by the UniMatch framework \cite{Yang-CVPR-2023}, the proposed method obtained {\it enhanced labels} with the improved quality of pseudo labels generated by SAM and CLIP.
We improve the accuracy of semantic segmentation by training the segmentation model so that the predictions of the model when strong perturbations are applied to unlabeled images correspond to their {\it enhanced labels}.
Through experiments using PASCAL VOC 2012 \cite{Everingham-IJCV-2015} and Microsoft COCO \cite{Lin-ECCV-2014}, we demonstrate the effectiveness of the proposed method compared to conventional methods.

\section{Related Work}
\label{sec:related work}

We give an overview of image segmentation, semi-supervised semantic segmentation, and applications combining SAM and CLIP.

\noindent
{\bf Image Segmentation} --- 
A lot of methods have been proposed for image segmentation since Fully Convolutional Network (FCN) \cite{Long-CVPR-2015} has been proposed.
Many techniques such as atrous convolution \cite{Chen-ECCV-2018} and pyramid pooling have been developed to achieve accurate image segmentation for a variety of images.
Recently, transformer-based methods have been proposed \cite{Zheng-CVPR-2021,Wang-ICCV-2021}.
SETR \cite{Zheng-CVPR-2021} employs Vision Transformer (ViT) \cite{Dosovitskiy-ICLR-2020} as the backbone of feature extraction, while PVT \cite{Wang-ICCV-2021} uses a transformer introducing a pyramid structure.
SegFormer \cite{Xie-NeurIPS-2021} consists of a hierarchical transformer encoder and a decoder with lightweight fully-connected layers.
Foundation models for image segmentation are also developed, such as Segment Anything Model (SAM) \cite{Kirillov-ICCV-2023}.
SAM is a segmentation model pre-trained on the SA-1B dataset of 11 million images annotated with over 1 billion labels, and achieves high generalization capability.
Recently, SAM has been used in combination with other foundation models to perform various tasks in zero-shot.

\noindent
{\bf Semi-Supervised Semantic Segmentation} ---
In many approaches for semi-supervised semantic segmentation, pseudo labels are assigned to unlabeled images based on the predictions of the model, and the model is trained using these pseudo labels as ground truth.
A method using the successive learning flow \cite{Huang-CVPR-2023} employs knowledge distillation, in which the teacher model creates pseudo labels for the student model.
One of the methods using the parallel learning flow is Cross Pseudo Supervision (CPS) \cite{Chen-CVPR-2021}.
CPS consists of mini batches of labeled and unlabeled data, and trains the two models so that their predictions are consistent.
UniMatch \cite{Yang-CVPR-2023} has been proposed to improve the accuracy of CPS.
UniMatch trains a single model so that its predictions are consistent when weak and strong perturbations are applied to the unlabeled image.
Perturbations are applied not only at the image level, such as cropping and color transformations, but also in the feature space.

\noindent
{\bf Applications of SAM and CLIP} ---
There are some studies \cite{Yu-BMVC-2023,Aleem-CVPR-2024,Wang-CVPR-2024} considering the combination of SAM \cite{Kirillov-ICCV-2023} and CLIP \cite{Radford-ICML-2021}.
Yu et al. \cite{Yu-BMVC-2023} proposed a method that combines SAM and CLIP for audio-visual segmentation \cite{Zhou-ECCV-2022}, which is a task to detect objects with sound emissions in video at pixel level.
Aleem et al. \cite{Aleem-CVPR-2024} proposed a medical image segmentation method, SaLIP, that combines SAM and CLIP.
Wang et al. \cite{Wang-CVPR-2024-2} proposed SAM-CLIP, which integrates SAM and CLIP into a single model using multi-task learning, continuous learning \cite{Li-TPAMI-2018}, and knowledge distillation.
On the other hand, to the best of our knowledge, there have been no studies using the combination of SAM and CLIP for semi-supervised learning.

\section{Proposed Method}
\label{sec:proposed method}

Semantic segmentation using semi-supervised learning generates pseudo labels from model predictions, and therefore the quality of the pseudo labels has a strong impact on model training.
We propose an image segmentation method that combines semi-supervised learning with pseudo-labels generated based on zero-shot annotation.
First, we introduce a zero-shot annotation method using SAM and CLIP to improve the quality of pseudo labels.
Next, pseudo labels are generated for unlabeled images using SAM and CLIP, and {\it enhanced labels} that improve the quality of these pseudo labels are generated using the semi-supervised learning framework of UniMatch \cite{Yang-CVPR-2023}.
Then, the segmentation model is trained so that the predictions of the model when strong perturbations are applied to unlabeled images correspond to their {\it enhanced labels}.
In the following, we describe zero-shot annotation using SAM and CLIP and enhanced label generation using the semi-supervised learning framework of UniMatch.

\subsection{Zero-Shot Annotation Using SAM and CLIP}
\label{sec:zero}

\begin{figure}[t]
  \centering
  \includegraphics[width=1.0\linewidth]{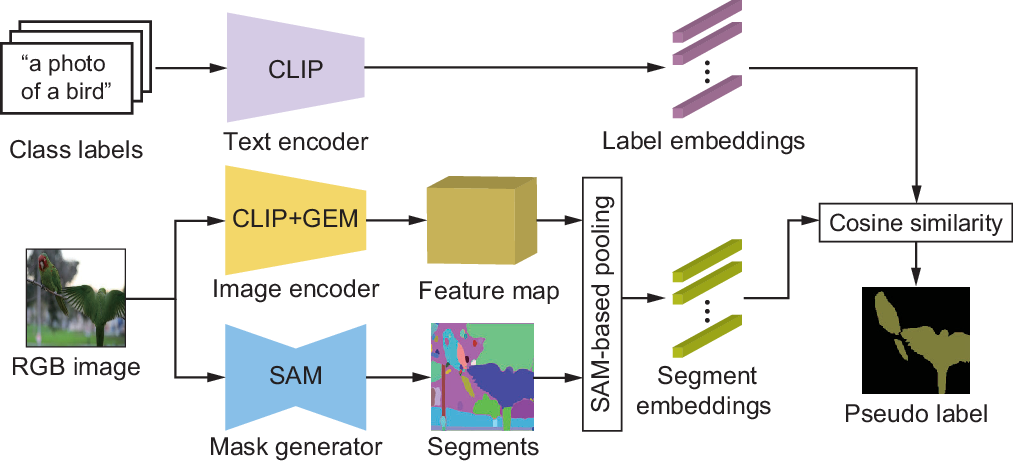} 
  \caption{Overview of zero-shot annotation using SAM and CLIP.} 
  \label{fig:sam_clip} 
\end{figure}
\begin{figure*}[t]
  \centering
  \includegraphics[width=.93\linewidth]{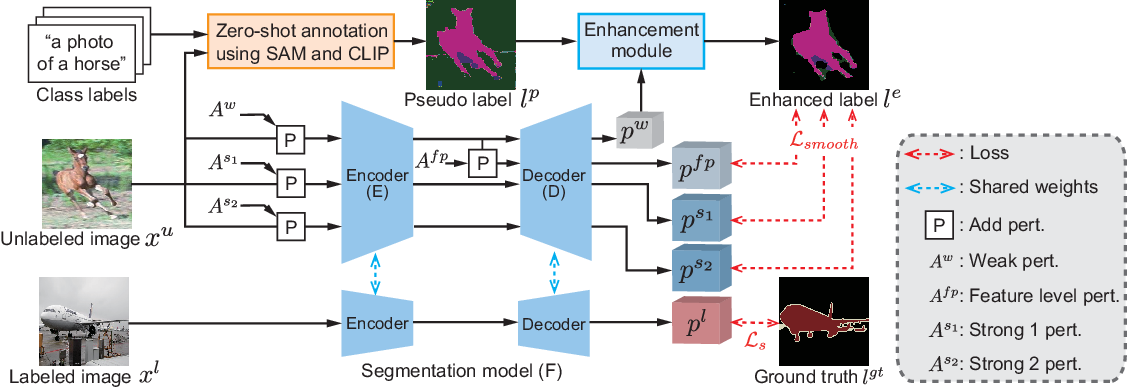} 
  \caption{Training flow of the proposed semi-supervised learning.} 
  \label{fig:proposed} 
\end{figure*}

We focus on zero-shot annotation for assigning pseudo labels independent of the predictions of the model to be trained in the semi-supervised learning framework.
It is necessary to divide the image into object-based segments and to assign class labels to the segments to achieve zero-shot annotation in image segmentation.
We employ SAM \cite{Kirillov-ICCV-2023}, which is a foundation model for image segmentation, to divide images into object-based segments.
Although SAM can be used to divide images into fine-grained segments, the released version of SAM does not assign class labels to each segment.
To address this problem, we employ CLIP \cite{Radford-ICML-2021}, which is a foundation model of vision and language, to assign a class label to each segment.
CLIP is a ViT-based model that can embed images and text in the same feature space, and consists of an image encoder for extracting embeddings from images and a text encoder for extracting embeddings from text.
The use of CLIP makes it possible to correspond a given class label to segments obtained by SAM.
Fig. \ref{fig:sam_clip} shows an overview of the zero-shot annotation using SAM and CLIP proposed in this paper.

CLIP does not take into account the position of objects in the image encoder since CLIP is trained to increase the similarity between the image and the text.
It is also observed that the embedding of patches containing objects corresponding to the text is similar to the embedding of patches around the object \cite{Bousselham-CVPR-2024}.
Therefore, we introduce the Grounding Everything Module (GEM) \cite{Bousselham-CVPR-2024} into the image encoder of CLIP.
GEM consists of self-self attention blocks that uses self-self attention as key-key, query-query, and value-value representations.
Self-self attention has a similar effect to clustering, making features from the same object similar while preserving consistency with the text embedding.

We obtain embeddings for each segment by performing SAM-based pooling on the feature map output from the image encoder of CLIP and the segments generated by SAM.
SAM-based pooling extracts segment embeddings from the feature map $I\in{\mathbb{R}^{H \times W \times D}}$ obtained from the image encoder of CLIP.
Let $S_k\in{\{0,1\}^{H \times W}}$ be the mask image for the $k$-th segment, the embedding $\bm{f}_k\in{\mathbb{R}^D}$ for $S_k$ is calculated by
\begin{equation}
  \bm{f}_k=\sum_{x,y}\frac{\bm{I}\odot{S_k}}{\sum_{x,y}{S_k}},
\end{equation}
where $\odot{}$ is the Adamar product, $(x,y)$ are the image coordinates, $D$ is the dimension of the CLIP feature space, and $1 \le x \le H$, $1 \le y \le W$.
We input class labels into the text encoder of CLIP and obtain the embedding for each class label, where the class label is the object name to be annotated, e.g., the class labels of the objects in the dataset.
Then, we calculate the cosine similarity $S_{c,k}$ between the embedding $\bm{T}_c\in{\mathbb{R}^D}$ for the $c$-th class label and the $k$-th segment embedding $\bm{f}_k\in{\mathbb{R}^D}$ by
\begin{equation}
  S_{c,k} = \frac{{\bm{f}_k}{\bm{T}_c}^T}{||\bm{f}_k||\cdot{||\bm{T}_c||}}.
\end{equation}
Finally, we obtain pseudo labels for the input images by assigning to each segment the class label that has the maximum similarity.
Note that if the cosine similarity for all class labels is not higher than a threshold, the segment is not assigned a class label, e.g., the background of the input image in Fig. \ref{fig:sam_clip}.

\subsection{Enhanced Label Generation}

We generate enhanced labels whose quality is improved based on the pseudo labels generated by SAM and CLIP inspired by the semi-supervised learning framework of UniMatch \cite{Yang-CVPR-2023}.
The flow of the semi-supervised learning proposed in this paper is illustrated in Fig. \ref{fig:proposed}.

The proposed method adds four types of perturbations $A^w$, $A^{fp}$, $A^{s_1}$, and $A^{s_2}$ to the unlabeled image $x^u \in \mathbb{R}^{H \times W \times 3}$ as in UniMatch \cite{Yang-CVPR-2023}, where $A^w$ indicates weak perturbations, $A^{fp}$ indicates perturbations on the feature space, and $A^{s_1}$ and $A^{s_2}$ indicate strong perturbations.
The outputs from the decoder when these perturbations are added are given by
\begin{eqnarray}
  p^{w} &=& F(A^w(x^u)),\\
  p^{fp} &=& D(A^{fp}(E(x^u))),\\
  p^{s_1} &=& F(A^{s_1}(x^u)),\\
  p^{s_2} &=& F(A^{s_2}(x^u)),
\end{eqnarray}
where $F$ indicates the segmentation model to be trained, $E$ represents the encoder of $F$, and $D$ represents the decoder of $F$.
For $x^u$, pseudo label $l^p$ is generated using zero-shot annotation using SAM and CLIP as described in Sect. \ref{sec:zero}.
We generate the enhanced label $l^e$ by inputting $l^p$ and $p^w$ into the enhancement module.
In the Enhancement module, the enhanced label $l^e$ is calculated by 
\begin{equation}
\begin{split}
  l^e =& \mathbb{1}(\MAX_c({p}^w)<\tau)\odot{l^p}\\
  &+\mathbb{1}(\MAX_c({p}^w)\geq \tau)\odot{\argmax_c {p}^w},
  \label{eq:le}
\end{split}
\end{equation}
where $\tau$ indicates the threshold for the confidence of $p^w$.
If the confidence of $p^w$ is lower than $\tau$ at a pixel, the pseudo label $l^p$ is adopted; otherwise, the estimated label of $p^w$ is adopted as the enhanced label $l^e$.

For $x_u$, the loss $\mathcal{L}_{smooth}$ is calculated between the enhanced label $l^e$ and $p^{fp}$, $p^{s_1}$, and $p^{s_2}$, respectively.
The loss $\mathcal{L}_{smooth}$ is the cross-entropy loss with label smoothing \cite{Szegedy-CVPR-2016}.
Label smoothing suppresses overfitting to a label by including ambiguity in the label, resulting in reducing errors in SAM and CLIP annotations.
In a one-hot vector representation, $1$ indicating the class is changed to $1-\epsilon$ and $0$ indicating the other classes is changed to $\epsilon / (C-1)$, where $\epsilon$ indicates a hyperparameter and $C$ is the number of classes.
In this paper, $\epsilon$ is the inverse of the number of classes in the dataset.
The loss $\mathcal{L}_{smooth}$ between the model output $p$ and the enhanced label $l^e$ is defined by
\begin{equation}
  \begin{split}
    \mathcal{L}_{smooth}(p)= -\frac{1}{N}&\sum_{i=1}^{N}\BiggParen{{(1-\epsilon)\log{p_{i}^{(l^e)_i^t}}}\\
    &+\sum_{c\in{C\backslash}\{(l^e)_i^t\}}{\frac{\epsilon}{C-1}\log{p_i^c}}},
  \end{split}
\end{equation}
where $N$ is the number of pixels, $C$ is a set of enhanced labels, $p_i^c$ is the prediction of the model for class $c$ at pixel $i$, and $(l^e)_i^t$ is the class at pixel $i$ of the enhanced label $l^e$.
Using the loss $\mathcal{L}_{smooth}$, the loss $\mathcal{L}_u$ for $x_u$ is given by
\begin{equation}
  \mathcal{L}_u = \mathcal{L}_{smooth}(p^{fp})+\mathcal{L}_{smooth}(p^{s_1})+\mathcal{L}_{smooth}(p^{s_2}).
\end{equation}
For labeled images $x^l \in \mathbb{R}^{H \times W \times 3}$, we compute the cross-entropy loss $\mathcal{L}_s$ between the ground truth label and the model output as in supervised learning.
The total loss used in the training of the proposed method is given by
\begin{equation}
  \mathcal{L} = \frac{1}{2}(\mathcal{L}_s+\mathcal{L}_u).
\end{equation}
In the proposed method, training is performed for each mini batch consisting of eight unlabeled images and eight labeled images.

\section{Experiments and Discussion}
\label{sec:experiments}

We evaluate the accuracy of the proposed method in semantic segmentation using public datasets to demonstrate the effectiveness of the proposed method.
We describe the experimental setup, an ablation study of the proposed method, and a comparison with existing methods in the following.

\subsection{Experimental Setup}

We describe the public datasets used in the experiments, the details of the implementation of the proposed method, and the evaluation metrics.

\noindent
{\bf Datasets} ---
In this experiment, we use two public datasets for training and evaluating image segmentation methods: the PASCAL VOC 2012 original (PASCAL)\footnote{\url{http://host.robots.ox.ac.uk/pascal/VOC/}} \cite{Everingham-IJCV-2015} and Microsoft COCO (COCO)\footnote{\url{https://cocodataset.org/}} \cite{Lin-ECCV-2014}.
PASCAL provides 1,464, 1,449, and 1,456 images for training, validation, and testing, respectively.
The images are labeled with 21 classes, including background.
To evaluate the accuracy of semantic segmentation in semi-supervised learning, we divide the training data into 1/16 (92), 1/8 (183), 1/4 (366), and 1/2 (732), as in the conventional methods \cite{Yang-CVPR-2023,Wang-CVPR-2024}, and conduct experiments by changing the number of labeled images, where the number in parentheses indicates the number of images.
COCO is a large dataset containing both indoor and outdoor scenes.
COCO provides 118,000 images for training and 5,000 images for testing, with 80 classes of object labels and a void label assigned to the images.
As in the conventional methods \cite{Yang-CVPR-2023,Wang-CVPR-2024}, we divide the training data into 1/512 (232), 1/256 (463), 1/128 (925), and 1/64 (1,849), and conduct experiments by changing the number of labeled images.

\noindent
{\bf Implementation Details} ---
As in the conventional methods \cite{Yang-CVPR-2023,Wang-CVPR-2024}, each mini batch consists of eight unlabeled images and eight labeled images.
The initial learning rates are 0.001 and 0.004 for PASCAL and COCO, respectively, and SGD is used as an optimizer.
The learning rate is updated by the poly learning rate scheduler.
As weak perturbation $A^w$, we use resize and crop with a probability of 100\%, and flip with a probability of 50\%.
As strong perturbation $A^s$, we use a combination of color transform and CutMix \cite{Yang-CVPR-2022}.
The color transform consists of Color Jitter with probability 0.8, grayscale transform with probability 0.2, and Gaussian Blur with probability 0.5.
The differences in the probabilities for color transform and the regions cut by CutMix result in differences between $A^{s_1}$ and $A^{S_2}$.
As perturbation $A^{fp}$ on the feature space, we use channel dropout with a probability of 50\%.
The threshold $\tau$ in Eq. (\ref{eq:le}) is set to 0.7 in this paper.
We input the prompt ``a photo of a \{classlabel\}'' to the text encoder of CLIP in the experiments.
In the experiments, SefFormer-B4 \cite{Xie-NeurIPS-2021} is used as the segmentation backbone in the proposed method, if not otherwise specified.
For the implementation environment, PyTorch 1.12.1 is used as the deep learning framework, and experiments are conducted on NVIDIA A100 GPU.

\noindent
{\bf Evaluation Metric} ---
In the experiments, we employ Intersection over Union (IoU) as the evaluation metric.
Let $Gt$ be the region of ground truth and $Pr$ be the predicted region of the model, the IoU for a class is calculated by
\begin{equation}
  \mathrm{IoU} = \frac{Gt \cap Pr}{Gt \cup Pr-Gt\cap Pr}.
\end{equation}
We calculate IoU for each class and use the average value, mIoU, as the evaluation metric.

\subsection{Ablation Study}

We first evaluate the performance of zero-shot annotation using SAM and CLIP and label smoothing to verify the effectiveness of the proposed method.

\subsubsection{Performance of Zero-Shot Annotation}

\begin{figure}[t]
  \centering
  \includegraphics[width=0.8\linewidth]{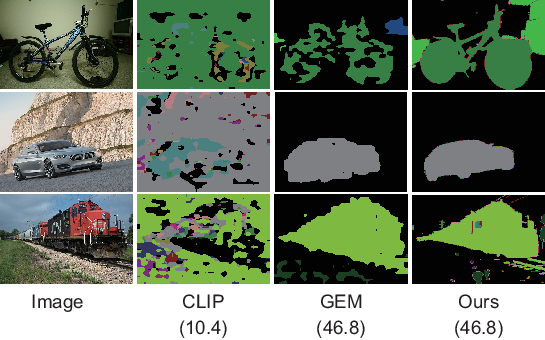} 
  \caption{Results of zero-shot annotation using SAM and CLIP for PASCAL.
  Values in parentheses indicate mIoU for each method.} 
  \label{fig:zero-shot}
\end{figure}

The performance of the zero-shot annotation method using SAM and CLIP proposed in this paper is compared with existing methods.
Note that although SAM-CLIP \cite{Wang-CVPR-2024-2} has been proposed as a method combining SAM and CLIP, this method is excluded from the comparison since it is fine-tuned using 40.8M images.
In this paper, we compare the proposed method with CLIP \cite{Radford-ICML-2021} and GEM \cite{Bousselham-CVPR-2024}, which are zero-shot segmentation methods without fine-tuning, since we assume application to semi-supervised learning, where the number of labeled images that can be used for training is limited.
Fig. \ref{fig:zero-shot} shows the result of zero-shot segmentation on the PASCAL dataset.
The proposed method and GEM \cite{Bousselham-CVPR-2024} have the highest mIoU, while the proposed method, unlike GEM, can generate masks that accurately identify the contour of the objects.

\subsubsection{Effect of Label Smoothing}

To demonstrate the effectiveness of label smoothing \cite{Szegedy-CVPR-2016}, we evaluate the accuracy of the proposed method with and without label smoothing.
We conduct the experiment on the Pascal dataset with 732 labeled images.
mIoU without label smoothing is 79.14, while mIoU with label smoothing is 80.7.
The above results demonstrate the effectiveness of label smoothing in the proposed method.

\subsection{Comparison with Existing Methods}
 
\begin{table}[t]
  \caption{Comparison with existing methods for PASCAL.
  Bold type indicates the best results for each splitting of the labeled images.}
  \label{tab:pascal}
  \centering
  \begin{widetable}{\linewidth}{lccccc}
    \toprule
    & \multicolumn{5}{c}{mIoU [\%] $\uparrow$} \\
    \cmidrule(lr){2-6}
    \multirow{2}{*}{Method}& 1/16  & 1/8 & 1/4 & 1/2 & Full \\
    & (92) & (183) & (366) & (732) & (1,464) \\
    \cmidrule(lr){1-6}
    UniMatch \cite{Yang-CVPR-2023} & 75.2 & 77.19 & 78.8 & 79.9 & 81.2 \\
    LogicDiag \cite{Liang-ICCV-2023} & 73.3 & 76.7 & 77.9 & 79.4 & ---\\
    AllSpark \cite{Wang-CVPR-2024} & 76.07 & 78.41 & 79.77 & 80.75 & 82.12 \\
    BeyondPixels \cite{Howlader-ECCV-2024} & {\bf 77.3 } & 78.6 & {\bf 79.8 } & {\bf 80.8 } & 81.7 \\
    Ours & 65.30 & {\bf 78.69} & {\bf 79.8 } & 80.56 & {\bf 82.15} \\
    \bottomrule
  \end{widetable}
\end{table}

To demonstrate the effectiveness of the proposed method, we compare its accuracy with UniMatch \cite{Yang-CVPR-2023}, LogicDiag \cite{Liang-ICCV-2023}, AllSpark \cite{Wang-CVPR-2024}, and BeyondPixels \cite{Howlader-ECCV-2024}, which are the state-of-the-art methods for semantic segmentation using semi-supervised learning.
Note that we compare semi-supervised semantic segmentation methods that propose a learning framework.
BeyondPixels \cite{Howlader-ECCV-2024} can be integrated into a semi-supervised learning framework, and therefore, in this experiment, we use BeyondPixels integrated into UniMatch \cite{Yang-CVPR-2023} as well as the proposed method.

Table \ref{tab:pascal} shows the experimental results for PASCAL.
The proposed method achieves higher mIoU than UniMatch \cite{Yang-CVPR-2023} and LogicDiag \cite{Liang-ICCV-2023} when the number of labeled images is greater than 183.
The proposed method achieves the same or higher accuracy compared to AllSpark \cite{Wang-CVPR-2024} and BeyondPixels \cite{Howlader-ECCV-2024}.
AllSpark uses images with $513 \times 513$ pixels for training, while the proposed method uses images with $321 \times 321$ pixels.
The proposed method can achieve semi-supervised learning comparable to AllSpark and BeyondPixels even with small image sizes.
Table \ref{tab:coco} shows the experimental results for COCO.
Note that there is no result for BeyondPixels \cite{Howlader-ECCV-2024} because of errors in the experiments for COCO even if the public code is used.
The proposed method outperforms UniMatch \cite{Yang-CVPR-2023}, LogicDiag \cite{Liang-ICCV-2023}, and AllSpark \cite{Wang-CVPR-2024} for all patterns of labeled images.
AllSpark and UniMatch train on images with $513 \times 513$ pixels, while the proposed method trains on images with $400 \times 400$ pixels.
Similar to PASCAL, the proposed method can perform semi-supervised learning in COCO even with small image sizes.

\begin{table}[t]
  \caption{Comparison with existing methods for COCO.
  Bold type indicates the best results for each splitting of the labeled images.}
  \label{tab:coco}
  \centering
  \begin{widetable}{\linewidth}{lcccc}
    \toprule
    & \multicolumn{4}{c}{mIoU [\%] $\uparrow$} \\
    \cmidrule(lr){2-5}
    \multirow{2}{*}{Method}& 1/512 & 1/256 & 1/128 & 1/64 \\
     & (232) & (463) & (925) & (1,849) \\
    \cmidrule(lr){1-5}
    UniMatch \cite{Yang-CVPR-2023} & 31.86 & 38.88 & 44.35 & 48.17 \\
    LogicDiag \cite{Liang-ICCV-2023} & 33.1 & 40.3 & 45.4 & 48.8 \\
    AllSpark \cite{Wang-CVPR-2024} & 34.10 & 41.65 & 45.48 & 49.56 \\
    Ours & {\bf 46.06} & {\bf 48.20} & {\bf 48.98} & {\bf 51.20} \\
    \bottomrule
  \end{widetable}
\end{table}

\begin{figure}[t]
  \centering
  \includegraphics[width=.78\linewidth]{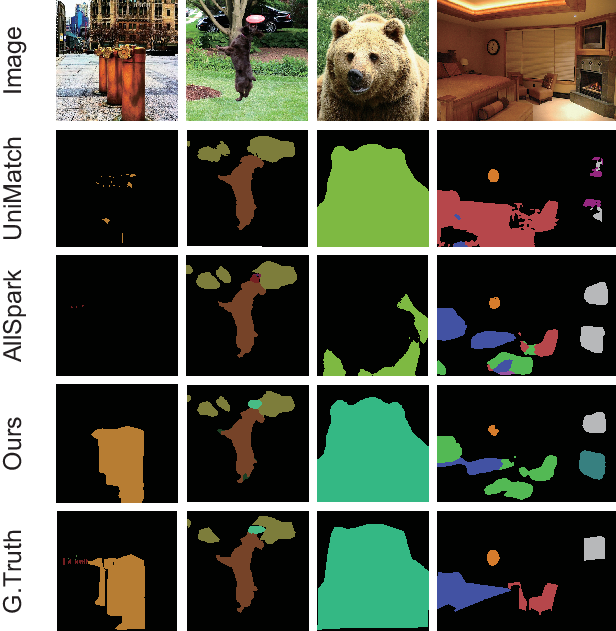}
  \caption{Semantic segmentation results of each method for COCO with 1/512 (232) split.}
  \label{fig:p_seg}
\end{figure}

Fig. \ref{fig:p_seg} shows the segmentation results of each method for COCO with 1/512 (232) split.
Since the images in COCO are labeled with 80 classes, some labels are not detected by UniMatch \cite{Yang-CVPR-2023} and AllSpark \cite{Wang-CVPR-2024}, which generate pseudo labels based on the prediction of the model to be trained.
On the other hand, the proposed method, which generates pseudo labels based on zero-shot annotation using SAM and CLIP, can detect most labels.
UniMatch and AllSpark have errors in the labels assigned to the detected segments, while the proposed method assigns the correct labels.
However, as shown in the right column of Fig. \ref{fig:p_seg}, the proposed method sometimes fails to assign labels correctly to small segments, requiring improvement in the accuracy of zero-shot annotation using SAM and CLIP.

\subsection{Comparison for Segmentation Backbones}

We compare the accuracy of each method when the segmentation backbone used is changed.
UniMatch \cite{Yang-CVPR-2023} uses DeepLabV3+ \cite{Chen-ECCV-2018}, which is a CNN-based segmentation backbone.
AllSpark \cite{Wang-CVPR-2024} is designed to use transformer-based segmentation backbones.
In this experiment, we compare the accuracy of R101+DeepLabV3+ \cite{Chen-ECCV-2018} used in UniMatch \cite{Yang-CVPR-2023} and SegFormer-B4 and B5 \cite{Xie-NeurIPS-2021} used in AllSpark \cite{Wang-CVPR-2024} as the segmentation backbone.
The experiment is conducted in PASCAL with 183 labeled images.
Table \ref{tab:backbone} shows the comparison of each method for segmentation backbones in PASCAL.
UniMatch exhibits the highest accuracy when DeepLabV3+ is used as the backbone.
AllSpark exhibits the highest accuracy when SegFormer is used, however, the accuracy significantly decreases when a CNN-based backbone is used.
On the other hand, the proposed method achieves high accuracy with any type of backbone.
In particular, the highest accuracy is achieved in all cases when SegFormer-B4 with a small number of parameters is used.

\begin{table}[t]
  \caption{Comparison of each method for segmentation backbones in PASCAL.
  Bold type indicates the best results for each}
  \label{tab:backbone}
  \centering
  \small
  \begin{widetable}{\linewidth}{lccc}
    \toprule
    \multirow{2}{*}{Backbone} & \multicolumn{3}{c}{mIoU [\%] $\uparrow$} \\
    \cmidrule(lr){2-4}
     & UniMatch \cite{Yang-CVPR-2023} & AllSpark \cite{Wang-CVPR-2024} & Ours \\
    \cmidrule(lr){1-4}
    R101+DeepLabV3+ \cite{Chen-ECCV-2018} & 77.19 & 73.70 & {\bf 77.65} \\
    SegFormer-B4 \cite{Xie-NeurIPS-2021} & 76.28 & 77.92 & {\bf 78.69} \\
    SegFormer-B5 \cite{Xie-NeurIPS-2021} & 76.56 & {\bf 78.41} & 78.16 \\
    \bottomrule
  \end{widetable}
\end{table}

\subsection{Comparison for Model Parameters}

We also discuss the model size of each method.
UniMatch \cite{Yang-CVPR-2023} and the proposed method are semi-supervised learning at the framework level, and therefore do not change the architecture of the model to be trained and do not increase the number of model parameters.
On the other hand, AllSpark \cite{Wang-CVPR-2024} is semi-supervised learning at the architecture level, and therefore changes the architecture of the model to be trained and increases the number of model parameters.
Comparing the proposed method and AllSpark, when training SegFormer-B5, the number of parameters in the proposed method is 84.7 M, while in AllSpark it is 89.4 M, resulting in an increase of 4.7 M in the number of parameters.
The proposed method is more efficient than AllSpark since the proposed method does not change the architecture of the model and does not increase the number of model parameters.

\section{Conclusion}
\label{sec:conclusion}

In this paper, we proposed a semi-supervised semantic segmentation method using zero-shot annotation with SAM \cite{Kirillov-ICCV-2023} and CLIP \cite{Radford-ICML-2021}.
We generate pseudo labels using zero-shot annotation with SAM and CLIP, and improve their quality by semi-supervised learning framework of UniMatch \cite{Yang-CVPR-2023} as {\it enhanced labels}.
Through experiments using PASCAL \cite{Everingham-IJCV-2015} and COCO \cite{Lin-ECCV-2014}, we demonstrated the effectiveness of the proposed method compared to the state-of-the-art semi-supervised learning methods: UniMatch \cite{Yang-CVPR-2023}, LogicDiag \cite{Liang-ICCV-2023}, AllSpark \cite{Wang-CVPR-2024}, and BeyondPixels \cite{Howlader-ECCV-2024}.
The proposed method can be trained on small-sized images and achieves high accuracy independent of the type of segmentation backbone.

\section{Acknowledgment}
This work was supported in part by JSPS KAKENHI 23H00463 and 25K03131.

% To start a new column (but not a new page) and help balance the last-page
% column length use \vfill\pagebreak.
% -------------------------------------------------------------------------
%\vfill
%\pagebreak
%\vfill\pagebreak

% References should be produced using the bibtex program from suitable
% BiBTeX files (here: strings, refs, manuals). The IEEEbib.bst bibliography
% style file from IEEE produces unsorted bibliography list.
% -------------------------------------------------------------------------
{\small
  \bibliographystyle{IEEEbib}
  \bibliography{strings,refs}
}

\end{document}